\documentclass[runningheads]{llncs}
\usepackage{bbding}

\usepackage{graphicx}
\usepackage{multirow}
\usepackage{booktabs}
\usepackage{diagbox}
\usepackage{bbm}
\usepackage{amsmath}
\begin{document}
%






\title{Scale Federated Learning for Label Set Mismatch in Medical Image Classification}

%
%

\author{Zhipeng Deng\inst{1} \and
Luyang Luo\inst{1}\and
Hao Chen\inst{1,2}\Envelope }

\authorrunning{Z. Deng et al.}

\institute{Department of Computer Science and Engineering, \and
Department of Chemical and Biological Engineering, \\
The Hong Kong University of Science and Technology, Hong Kong, China
\\
\email{zdengaj@connect.ust.hk}, \email{cseluyang@ust.hk}, \email{jhc@cse.ust.hk}}
\titlerunning{FedLSM} 

\maketitle              

\begin{abstract}
Federated learning (FL) has been introduced to the healthcare domain as a decentralized learning paradigm that allows multiple parties to train a model collaboratively without privacy leakage. However, most previous studies have assumed that every client holds an identical label set. In reality, medical specialists tend to annotate only diseases within their area of expertise or interest. This implies that label sets in each client can be different and even disjoint. In this paper, we propose the framework FedLSM to solve the problem of \textbf{L}abel \textbf{S}et \textbf{M}ismatch. FedLSM adopts different training strategies on data with different uncertainty levels to efficiently utilize unlabeled or partially labeled data as well as class-wise adaptive aggregation in the classification layer to avoid inaccurate aggregation when clients have missing labels. We evaluated FedLSM on two public real-world medical image datasets, including chest X-ray (CXR) diagnosis with 112,120 CXR images and skin lesion diagnosis with 10,015 dermoscopy images, and showed that it significantly outperformed other state-of-the-art FL algorithms. The code can be found at https://github.com/dzp2095/FedLSM.

\end{abstract}
\section{Introduction}
Federated learning (FL) \cite{mcmahan2017communication} is an emerging decentralized learning paradigm that enables multiple parties to collaboratively train a model without sharing private data. FL was initially developed for edge devices, and it has been extended to medical image analysis to protect clinical data \cite{dong2022federated,jiang2022dynamic,liu2021federated,xu2022closing}. Non-identically independently distributed (Non-IID) data among clients is one of the most frequently stated problems with FL \cite{jiang2022harmofl,li2021model,li2020federated,li2021fedbn}. However, most studies of non-IID FL assumed that each client owns an identical label set, which does not reflect real-world scenarios where classes of interest could vary among clients. In the medical field, for instance, datasets from different centers (\emph{e.g.}, hospitals) are generally annotated based on their respective domains or interests. As a result, the label sets of different centers can be non-identical, which we refer to as \textbf{label set mismatch}.

To this end, we propose to solve this challenging yet common scenario where each client holds different or even disjoint annotated label sets. We consider not only single-label classification but also multi-label classification where partial labels exist, making this problem setting more general and challenging. Specifically, each client has data of locally identified classes and locally unknown classes. Although locally identified classes differ among clients, the union of identified classes in all clients covers locally unknown classes in each client. 

\begin{figure}[t!]
\centering
\includegraphics[width=0.9\textwidth]{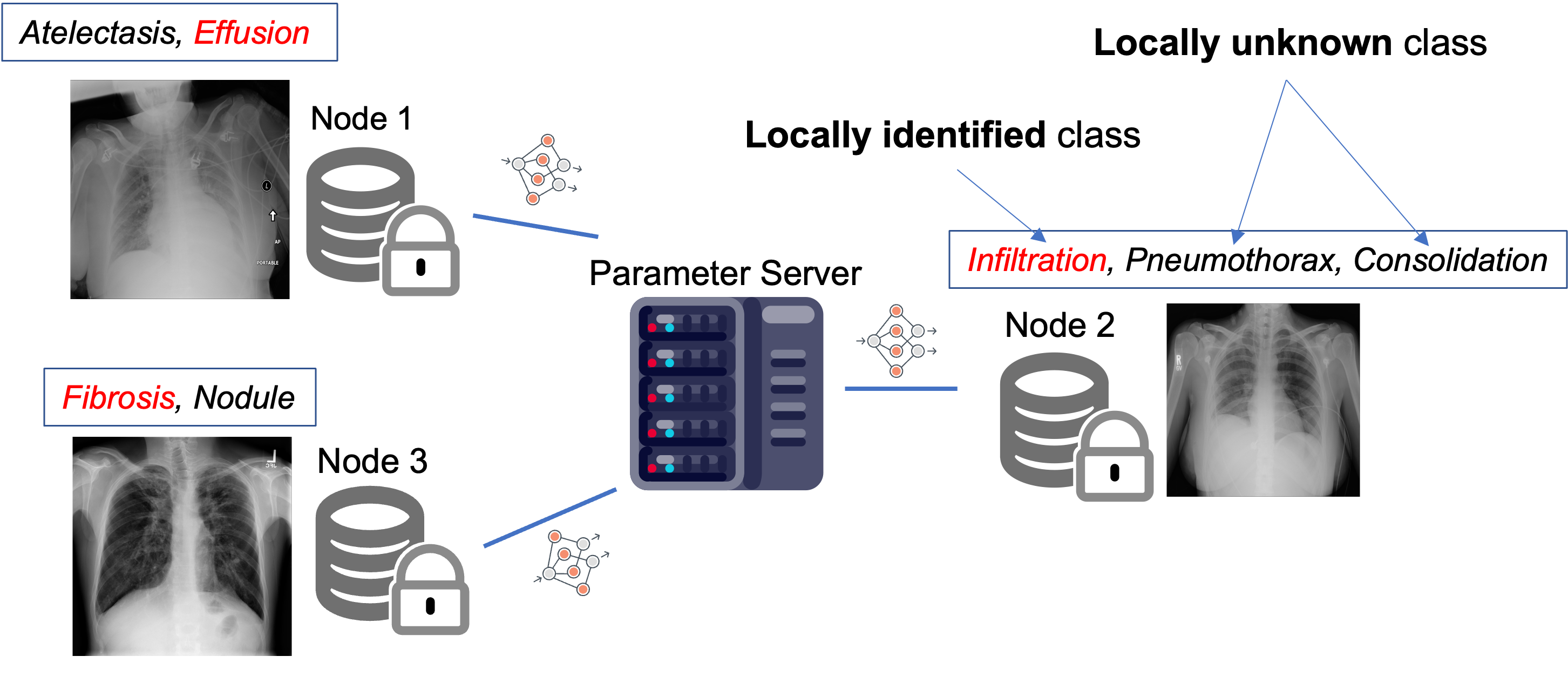}
\caption{Illustration of the \textbf{Label Set Mismatch} problem. Each node (client) has its own identified class set, which can differ from other nodes. The label in the box represents the correct \textbf{complete label}, while the label in the red text represents the \textbf{partial label} actually assigned to the image.} \label{fig1}
\end{figure}

There are few studies directly related to the label set mismatch scenario. The previous attempts were either limited in scope or have not achieved satisfactory results. For instance, FedRS \cite{li2021fedrs} assumed that each client only owns locally identified classes. FedPU \cite{lin2022federated} assumed that each client owns labels of locally identified classes and unlabeled data of all classes but it was not applicable to multi-label classification. FPSL \cite{dong2022federated} was designed for federated partially supervised learning which only targets multi-label classification. Over and above that, FedRS and FedPU tried to solve this problem only through local updating and ignored the server aggregation process in FL, leading to unsatisfactory performance. FPSL used bi-level optimization in the local training, which is only effective when the data is very limited. Federated semi-supervised learning (FedSemi) \cite{bdair2021fedperl,jeong2020federated,jiang2022dynamic,liu2021federated,yang2021federated} is another related field, but almost all of them assumed that some annotated clients \cite{bdair2021fedperl,jeong2020federated,liu2021federated} or a server \cite{jiang2022dynamic,jeong2020federated,yang2021federated} own labels of all classes. However, in real-world scenarios, especially in the healthcare domain, each client may only annotate data of specific classes within their domains and interests.

In this paper, we present FedLSM, a framework that aims to solve \textbf{L}abel \textbf{S}et \textbf{M}ismatch and is designed for both single-label and multi-label classification tasks. FedLSM relies on pseudo labeling on unlabeled or partially labeled samples, but pseudo labeling methods could lead to incorrect pseudo labels and ignorance of samples with relatively lower confidence. To address these issues, we evaluate the uncertainty of each sample using entropy and conduct pseudo labeling only on data with relatively lower uncertainty. We also apply MixUp\cite{zhang2017mixup} between data with low and high uncertainty and propose an adaptive weighted averaging for the classification layer that considers the client class-wise data numbers. We validated our propose method on two real-world tasks, including Chest X-ray (CXR) \cite{luo2020deep,luo2021oxnet} diagnosis (multi-label classification) and skin lesion diagnosis \cite{bdair2021fedperl} (single-label classification). Extensive experiments demonstrate that our method outperforms a number of state-of-the-art FL methods, holding promise in tackling the label set mismatch problem under federated learning.

\section{Methodology}
\subsection{Problem Setting}

\begin{figure}[htbp]
\centering
\includegraphics[width=0.95\textwidth]{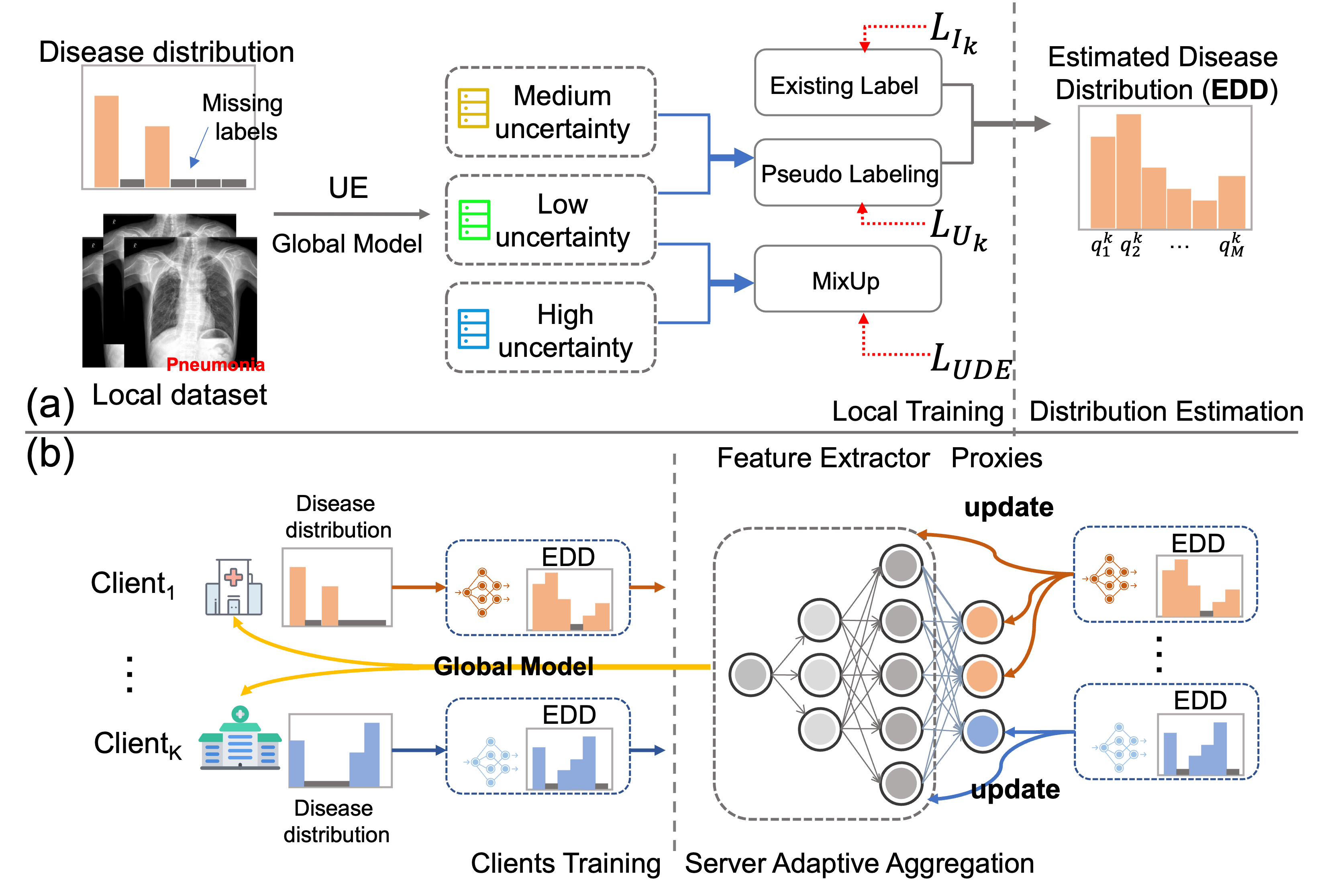}
\caption{Overview of the FedLSM framework. (a) Client training details, including uncertainty estimation (UE) and different training strategies for data with different uncertainty levels. The estimated disease distribution (EDD) is calculated using existing labels and pseudo labels. (b) Overview of the proposed scenario and FL paradigm, where each client has non-identical missing labels. EDD represents each client's contribution to each proxy in the adaptive classification layer aggregation.} \label{fig2}
\end{figure}

We followed the common FL scenario, where there are \emph{K} clients and one central server. Each client owns a locally-identified class set $\mathcal{I}_k$ and a locally-unknown class set $\mathcal{U}_k$. Although $\mathcal{I}_k$ and $\mathcal{U}_k$ can vary among clients and may even be disjoint, all clients share an identical class set as: 

\begin{equation}
    \mathcal{C} = \mathcal{I}_k \cup \mathcal{U}_k, \   k = 1,\  2,\  ...,\  K.
\end{equation}

Despite the fact that each client only identifies a subset of the class set $\mathcal{C}$, the union of locally identified class sets equals $\mathcal{C}$, which can be formulated as:
\begin{equation}
    \mathcal{C} = \mathcal{I}_1 \cup \mathcal{I}_2 \ ...\  \cup \mathcal{I}_{K}.
\end{equation}

The local dataset of client $\mathit{k}$ is denoted as $\mathcal{D}_k=\{(x^i,y^i)\}_{i=1}^{n_{k}}$, where $n_{k}$ denotes the number of data, $x_i$ is the $\mathit{i}$-th input image, and $y^i=[y^i_1,\ ...,\ y^i_c,\ ..., \ y^i_{M}]$ is the $\mathit{i}$-th label vector, M is the number of classes in $\mathcal{C}$ and $y^i_c$ refers to the label of class $\mathit{c}$. Notably, if $c \in \mathcal{I}_k$, $y^i_c \in \{0, 1\}$. If $c \in \mathcal{U}_k$, $y^i_c$ is set to 0.

 For subsequent illustration, we denote the backbone model as $f(\cdot)=f_{\Psi}(f_{\theta}(\cdot))$, where $f_{\theta}(\cdot)$ refers to the feature extractor with parameters $\theta$ and $f_{\Psi}$ refers to the last classification layer with parameters $\Psi = \{{\psi_c}\}_{c=1}^{M}$. Adopting the terminology from previous studies \cite{li2021fedrs,snell2017prototypical}, we refer to $\Psi$ as \emph{proxies} and $\psi_c$ as \emph{c}-th \emph{proxy}.

\subsection{Overview of FedLSM}

Our proposed framework is presented in Fig.~\ref{fig2}. As depicted in Fig.~\ref{fig2} (a), we evaluate the uncertainty of each sample in the local dataset $\mathcal{D}_k$ using the global model $f(\cdot)$ and split it into three subsets based on the uncertainty level. The low and medium uncertainty subsets are used for pseudo labeling-based training, while the low and high uncertainty subsets are combined by MixUp \cite{zhang2017mixup} to efficiently utilize uncertain data that might be ignored in pseudo labeling. The estimated disease distribution $q^k_c$ on client $k$ is calculated using the existing labels and pseudo labels. After local training, each client sends its estimated disease distribution $q^k$ and model weight $f_{k}(\cdot)$ to the central server. The feature extractors $\Theta$ are aggregated using FedAvg \cite{mcmahan2017communication} while proxies $\Psi$ are aggregated using our proposed adaptive weighted averaging with the help of $q^k$.

\subsection{Local Model Training}
\textbf{Uncertainty Estimation (UE).} We use the global model to evaluate the uncertainty of each sample in the local dataset by calculating its entropy \cite{robinson2008entropy}. The calculation of entropy in single-label classification is defined as:

\begin{equation}
H(x^{i})=-\sum_{c}{P(y=c|x^{i})log(P(y=c|x^{i}))}
\end{equation}

\noindent where $P(y=c|x^{i})$ refers to the predicted probability for a given class c and input $x^{i}$. The calculation of entropy in multi-label classification is similar to single-label classification where we only consider $\mathcal{U}_k$ and normalize the result to [0, 1]. After calculating the entropy, we empirically determine $n_{k}^{h}$ and $n_{k}^{l}$ and group the data with top-$n_{k}^{h}$ \textbf{h}ighest entropy as uncertain dataset $\mathcal{D}_k^{h}$, top-$n_{k}^{l}$ \textbf{l}owest entropy as confident dataset $\mathcal{D}_k^{l}$, and the rest as $\mathcal{D}_k^{m}$.

\noindent \textbf{Pseudo Labeling.} After local models are trained and aggregated on the central server, the resulting global model can identify the entire class set $\mathcal{C}$. Thereafter, we can use pseudo labeling-based method to leverage partially labeled or unlabeled data in $\mathcal{D}_k^{l}\cup \mathcal{D}_k^{m}$. We use the weakly-augmented version (i.e., slightly modified via rotations, shifts, or flips) of $x^i$ to generate pseudo labels on locally unknown classes by the teacher model. The loss $\mathcal{L}_{\mathcal{I}_k}$ applied on locally identified class set $\mathcal{I}_k$ is cross-entropy, and the loss applied on the locally unknown class set $\mathcal{U}_k$ of \emph{k}-th client in single-label classification can be formulated as:

\begin{equation}
\mathcal{L}_{\mathcal{U}_k} = -\frac{1}{N_{\mathcal{U}_k}}\sum_{i=1}^{N_{\mathcal{U}_k}}\sum_{c\in \mathcal{U}_k} \mathbbm{1}(f_{g,c}(\alpha(x^i)) \ge \tau)  {\hat{y}_c^{i}\mathrm{log}f_{k,c}(\mathcal{A}(x)^{i})}
\end{equation}

\noindent where $N_{\mathcal{U}_k}$ denotes the number of unlabeled data, $f_{k,c}(\cdot)$ is the predicted probability for class $\mathit{c}$ of the student model on \emph{k}-th client,  $\alpha(\cdot)$ and $\mathcal{A}(\cdot)$ refer to weak and strong augmentation respectively, $f_{g,c}(\alpha(x^i))$ is the predicted probability of class \emph{c} on the weakly augmented version of $x^i$ by the teacher model, $\hat{y}^{i}=\mathrm{arg max}(f_{g,c}(\alpha(x^i))$ is the pseudo label and $\tau$ is the threshold used to filter unconfident pseudo label. Specifically, the teacher model and student model both are initialized from the global model, while the teacher model is updated using exponential moving average (EMA) with the weights of the student model during training. Likewise, the loss function $\mathcal{L}_{\mathcal{I}_k}$ in multi-label classification is binary cross entropy and $\mathcal{L}_{\mathcal{U}_k}$ is:


\begin{equation}
\begin{split}
\mathcal{L}_{\mathcal{U}_k} =&  -\frac{1}{N_{\mathcal{C}}}\sum_{i=1}^{N_{\mathcal{C}}} \sum_{c\in \mathcal{U}_k} [ \mathbbm{1}(f_{g,c}(\alpha(x^i)) \ge \tau_p) \hat{y}^i_c\mathrm{log}(f_{k,c}(\mathcal{A}(x^{i}))) \\
& + \mathbbm{1}(f_{g,c}(\alpha(x^i)) \le \tau_n)(1-\hat{y}^i_c)\mathrm{log}(1-f_{k,c}(\mathcal{A}(x^{i}))) ]
\end{split}
\end{equation}
\noindent where $\tau_p$ and $\tau_n$ are the confidence threshold for positive and negative labels, $N_{\mathcal{C}}$ is the number of data.

\noindent \textbf{Uncertain Data Enhancing (UDE).} The pseudo label filtering mechanism makes it difficult to acquire pseudo-labels for uncertain data, which results in their inability to contribute to the training process. To overcome this limitation, we propose to MixUp \cite{zhang2017mixup} dataset with \textbf{l}owest entropy (confident) $\mathcal{D}_k^{l}$ and dataset with \textbf{h}ighest entropy as (uncertain) $\mathcal{D}_k^{h}$ to generate softer label $\Tilde{y}$ and input $\Tilde{x}$ as \begin{equation}
\begin{split}
\Tilde{x} =& \lambda x_l + (1-\lambda)x_h \\
\Tilde{y} =& \lambda y_l +  (1-\lambda)y_h
\end{split}
\end{equation}, where $x_l \in \mathcal{D}_k^{l}$ and $x_h \in \mathcal{D}_k^{h}$, and $y_l$ and $y_h$ are their corresponding labels or pseudo labels, respectively. We generate pseudo labels for uncertain data $(x_h,y_h)$ with a relatively smaller confidence threshold. The UDE loss function $\mathcal{L}_{UDE}$ is cross-entropy loss.

\noindent \textbf{Overall loss function.} The complete loss function is defined as: $\mathcal{L} = \mathcal{L}_{\mathcal{I}_k} + \mathcal{L}_{\mathcal{U}_k} + \lambda \mathcal{L}_{UDE} $, where $\lambda$ is a hyperparameter to balance different objectives.

\subsection{Server Model Aggregation}
After local training, the server will collect all the client models and aggregate them into a global model. In the \emph{r}-th round, the aggregation of the feature extractors $\{\theta_k\}_{k=1}^K$ is given by: $\theta^{r+1} \leftarrow \sum_{k}^{K}{\frac{n_k}{n}\theta_{k}^{r}}$, where $n=\sum_{k}^{K}n_k$. 

\noindent \textbf{Adaptive weighted proxy aggregation (AWPA).} As analyzed in \cite{li2021fedrs}, due to the missing labels of locally unknown class set $\mathcal{U}_k$ on \emph{k}-th client, the corresponding proxies $\{{\psi_{k,c}}\}_{c\in \mathcal{U}_k}$ are inaccurate and will further cause error accumulation during model aggregation. FedRS \cite{li2021fedrs} and FedPU \cite{lin2022federated} both seek to solve this problem only through local training while we use pseudo labels and the existing labels to indicate the contribution of aggregation of proxies as:  
\begin{equation}
    \psi_{c}^{r+1} \leftarrow \sum_{k=1}^{K}\frac{q_c^k}{\sum_{j=1}^{K}q_c^j}{\psi^r_{k,c}}
\end{equation}
\noindent where $q_c^k$ refers to the number of training data of the \emph{c}-th class on the \emph{k}-th client. During training, if $c\in \mathcal{U}_k$, $q_c^k$ is estimated by the number of pseudo labels as $q_c^k=\sum_{i=1}^{n_k}\hat{y}^i_c$. The weighting number of each client is modulated in an adaptive way through the pseudo labeling process in each round. 

\section{Experiments}
\subsection{Datasets}

We evaluated our method on two real-world medical image classification tasks, including the CXR diagnosis (multi-label) and skin lesion diagnosis (single-label).

\textbf{Task 1) NIH CXR diagnosis}. We conducted CXR diagnosis with NIH ChestX-ray14 dataset \cite{wang2017chestx}, which contains 112,120 frontal-view CXR images from 32,717 patients. NIH CXR diagnosis is a multi-label classification task and each image is annotated with 14 possible abnormalities (positive or negative). 

\textbf{Task 2) ISIC2018 skin lesion diagnosis}. We conducted skin lesion diagnosis with HAM10000 \cite{tschandl2018ham10000}, which contains 10,015 dermoscopy images. ISIC2018 skin lesion diagnosis is a single-label multi-class classification task where seven exclusive skin lesion sub-types are considered. 

Training, validation and testing sets for both datasets were divided into 7:1:2.

\subsection{Experiment Setup}
\textbf{FL setting.} We randomly divided the training set into \emph{k} client training sets and randomly select \emph{s} classes as locally identified classes on each client. We set the number of clients $k=8$, the number of classes $s=3$ for Task 1 and $k=5, s=3$ for Task 2. Please find the detail of the datasets in the supplementary materials.

\begin{table}[t!]
\caption{Comparison with state-of-the-art methods on NIH CXR diagnosis. \\ (\textbf{a}) FedAvg with 100\% labeled data (\textbf{b}) FedAvg \cite{mcmahan2017communication} (\textbf{c}) FedAvg* \cite{mcmahan2017communication} (\textbf{d}) FedProx* \cite{li2020federated} (\textbf{e}) MOON* \cite{li2021model} (\textbf{f}) FedRS \cite{li2021fedrs} (\textbf{g}) FPSL \cite{dong2022federated} (\textbf{h}) FedAvg-FixMatch* \cite{sohn2020fixmatch} (\textbf{i}) FSSL* \cite{yang2021federated}. * denotes the use of \textbf{task-dependent model aggregation} in FPSL\cite{dong2022federated}.} \label{tab:1}
\centering
\begin{tabular}{@{}c|c|cccc|cc|cc|c@{}}
\toprule

\diagbox{Disease}{Methods} & (a) & (b) & (c) & (d) & (e) & (f) & (g) & (h) & (i) & Ours \\ \midrule
Consolidation & 0.750 & 0.682 & 0.724 & 0.716 & 0.718 & 0.685 & 0.709 & \textbf{0.730} & 0.713 & 0.727 \\
Pneumonia & 0.717 & 0.695 & 0.684 & \textbf{0.707} & 0.700 & 0.685 & 0.711 & 0.699 & 0.706 & 0.700 \\
Effusion & 0.828 & 0.710 & 0.750 & 0.766 & 0.760 & 0.732 & 0.773 &0.800 & 0.759 & \textbf{0.804} \\
Emphysema & 0.913 & 0.717 & 0.842 & 0.842 & 0.877 & 0.644 & 0.837 &0.813 & 0.705 & \textbf{0.901} \\
Edema & 0.849 & 0.765 & 0.814 & 0.811 & 0.817 & 0.756 & 0.813 & 0.815 & 0.805 & \textbf{0.833} \\
Atelectasis & 0.776 & 0.634 & 0.721 & 0.716 & 0.722 & 0.660 & 0.712 &0.741 & 0.682 & \textbf{0.757} \\
Nodule & 0.768 & 0.707 & 0.722 & 0.729 & 0.720 & 0.685 & 0.716 & 0.735 & 0.695 & \textbf{0.742} \\
Mass & 0.820 & 0.686 & 0.725 & 0.723 & 0.736 & 0.670 & 0.704 & 0.767 &0.682 & \textbf{0.775} \\
Thickening & 0.775 & 0.726 & 0.732 & 0.737 & 0.740 & 0.720 & 0.725 &0.745 & 0.726 & \textbf{0.755} \\
Cardiomegaly & 0.877 & 0.750 & 0.830 & 0.840 & 0.841 & 0.707 & 0.820 & \textbf{0.858} & 0.794 & 0.846 \\
Fibrosis & 0.817 & 0.760 & 0.766 & 0.772 & 0.780 & 0.750 & 0.775 &0.776 & 0.790 & \textbf{0.804} \\
Hernia & 0.850 & 0.623 & 0.840 & \textbf{0.886} & 0.873 &0.609 & 0.870 & 0.865 & 0.837 & 0.884 \\
Pneumothorax & 0.865 & 0.790 & 0.818 & 0.805 & 0.836 & 0.761 & 0.809 & 0.836 & 0.766 & \textbf{0.858} \\
Infiltration & 0.696 & 0.695 & 0.687 & 0.678 & 0.689 & 0.682 & 0.672 & 0.680 & \textbf{0.700} & 0.680 \\ \midrule
Average AUC & 0.807 & 0.710 & 0.760 & 0.766 & 0.772 & 0.697 & 0.760 & 0.776 & 0.740 & \textbf{0.791} \\ \bottomrule
\end{tabular}
\end{table}

\noindent \textbf{Data augmentation and preprocessing.} The images in Task 1 were resized to 320$\times$320, while in Task 2, they were resized to 224$\times$224. For all experiments, weak augmentation refered to horizontal flip, and strong augmentation included a combination of random flip, rotation,  translation, scaling and one of the blur transformations in gaussian blur, gaussian noise and median blur.

\noindent \textbf{Evaluation Metrics.} For Task 1, we adopted AUC to evaluate the performance of each disease. For task 2, we reported macro average of AUC, Accuracy, F1, Precision and Recall of each disease. All the results are averaged over 3 runs.

\noindent \textbf{Implementation Details.} We used DenseNet121\cite{huang2017densely} as the backbone for all the tasks. The network was optimized by Adam optimizer where the momentum terms were set to 0.9 and 0.99. The total batch size was 64 with 4 generated samples using UDE. In task 1, we used the weighted binary cross-entropy as in FPSL\cite{dong2022federated}.  The local training iterations were 200 and 30 for Task 1 and Task 2, respectively, while the total communication rounds were 50 for both tasks. Please find more detailed hyperparameter settings in the supplementary material.



\begin{table}[htb]
\centering
\caption{Comparison with state-of-the-art methods on ISIC2018 Skin Lesion diagnosis.* denotes the use of \textbf{task-dependent model aggregation} in FPSL\cite{dong2022federated}.}\label{tab:2}
\begin{tabular}{@{}cccccc@{}}
\toprule
Methods & AUC & Accuracy & F1 & Precision & Recall \\ \midrule
\begin{tabular}[c]{@{}c@{}}FedAvg with 100\% labeled data\end{tabular} & 0.977 & 0.889 & 0.809 & 0.743 & 0.800 \\ \midrule
FedAvg \cite{mcmahan2017communication} & 0.927 & 0.810 & 0.576 & 0.721 & 0.622 \\
FedAvg* \cite{mcmahan2017communication} & 0.949 & 0.817 & 0.620 & 0.753 & 0.597 \\
FedProx* \cite{li2020federated} & 0.952 & 0.820 & 0.630 & \textbf{0.768} & 0.612 \\
MOON* \cite{li2021model} & 0.948 & 0.826 & 0.652 & 0.755 & 0.620 \\ \midrule
FedPU \cite{lin2022federated} & 0.927 & 0.796 & 0.550 & 0.699 & 0.570 \\
FedRS \cite{li2021fedrs} & 0.926 & 0.800 & 0.577 & 0.716 & 0.597 \\ 
FPSL \cite{dong2022federated} & 0.952 & 0.825 & 0.638 & 0.728 & 0.613 \\ \midrule
FedAvg* + FixMatch \cite{sohn2020fixmatch} & 0.940 & 0.789 & 0.564 & 0.681 & 0.541 \\
FSSL* \cite{yang2021federated} & 0.939 & 0.807 & 0.608 & 0.740 & 0.580 \\ \midrule
Ours & \textbf{0.960} & \textbf{0.846} & \textbf{0.713} & 0.763 & \textbf{0.699} \\ \bottomrule
\end{tabular}
\end{table}

\begin{table}[htb]
\centering
\caption{Ablation studies in terms of major components on task 1 and task 2.* denotes the use of \textit{masked aggregation} of proxies.}\label{tab:3}
\begin{tabular}{@{}c|c|ccccc@{}}
\toprule
\multirow{2}{*}{Method}& Task 1 & \multicolumn{5}{c}{Task 2} \\ \cmidrule{2-7}
  & AUC & AUC & Accuracy & F1 & Precision & Recall \\ \midrule
Ours* w/o AWPA & 0.784 & 0.950 & 0.837 & 0.676 & \textbf{0.783} & 0.646 \\
Ours w/o UDE & 0.689 & 0.954 & 0.825 & 0.655 & 0.774 & 0.638 \\
Ours w/o UE & 0.705 & 0.951 & 0.832 & 0.660 & 0.697 & 0.646 \\
Ours & \textbf{0.791} & \textbf{0.960} & \textbf{0.846} & \textbf{0.713} & 0.763 & \textbf{0.699} \\
\bottomrule 
\end{tabular}
\end{table}

\subsection{Comparison with state-of-the-arts and ablation study}

We compared our method with recent state-of-the-art (SOTA) non-IID FL methods, including FedProx \cite{li2020federated}, which applied $L_2$ regularization, and MOON \cite{li2021model}, which introduced contrastive learning. We also compared with other SOTA non-IID FL methods that shared a similar setting with ours, including FedRS \cite{li2021fedrs}, which restricted proxy updates of missing classes, FedPU \cite{lin2022federated}, which added a misclassification loss, and FPSL \cite{dong2022federated}, which adopted task-dependent model aggregation. Additionally, we compared with FedSemi methods that can be easily translated into the label set mismatch scenario including FSSL \cite{yang2021federated} and FedAvg with FixMatch \cite{sohn2020fixmatch}. For our evaluation, we used FedAvg with 100\% labeled data as the benchmark and FedAvg trained with the same setting as the lower bound.

The quantitative results for the two tasks are presented in Table~\ref{tab:1} and Table~\ref{tab:2}. To ensure a fair comparison, we adopted the task-dependent model aggregation proposed in FPSL \cite{dong2022federated} in most of the compared FL methods, with the exception of FedRS and FedPU which are specifically designed for the similar scenario with us. Our proposed FedLSM achieves the best performance on almost all metrics. Notably, the improvement over the second-best method is 1.5\% for average AUC on Task 1 and 6.1\% for F1-score on Task 2.

\noindent \textbf{Ablation study.} We conducted ablation studies to assess the effectiveness of the primary components of our FedLSM framework. As depicted in Table~\ref{tab:3}, the performance drops significantly without UE or UDE. On the other hand, the adoption of AWPA boosts the performance by 0.7\% in AUC for Task 1, 3.7\% in F1-score, and 5.3\% in recall for Task 2.

\section{Conclusion}

We present an effective framework FedLSM to tackle the issue of label set mismatch in FL. To alleviate the impact of missing labels, we leverage uncertainty estimation to partition the data into distinct uncertainty levels. Then, we apply pseudo labeling to confident data and uncertain data enhancing to uncertain data. In the server aggregation phase, we use adaptive weighted proxies averaging on the classification layer, where averaging weights are dynamically adjusted every round. Our FedLSM demonstrates notable effectiveness in both CXR diagnosis (multilabel classification) and ISIC2018 skin lesion diagnosis (single-label classification) tasks, holding promise in tackling the label set mismatch problem under federated learning.

\subsubsection{Acknowledgement}

This work was supported by the Hong Kong Innovation and Technology Fund (Project No. ITS/028/21FP) and Shenzhen Science and Technology Innovation Committee Fund (Project No. SGDX20210823103201011).

%
%
%
\bibliographystyle{splncs04}
\bibliography{reference}

\newpage
\section{Supplementary Materials}
\begin{table}[]
  \centering
\caption{Hyperparameters setting of both tasks.} \label{tab:2}  \begin{tabular}{@{}ccccccccccccccc@{}}
    \toprule
     & $\tau$ & $\tau_{l}$ & $\tau_p$ & $\tau_n$ & $\tau_{lp}$ & $\tau_{ln}$ & $\lambda$ & EMA & $\alpha$ in MixUp & lr & lr\_decay   \\ \midrule
    Task 1  & /    & /    & 0.85& $5e^{-3}$ & 0.7 & $1e^{-2}$ &  0.1 & 0.999 & 0.2 & $1e^{-4}$ & $5e^{-4}$\\
    Task 2  & 0.95 & 0.85 & /   & /       & /   & / &  0.1 & 0.999 & 0.2 & $1e^{-4}$ & $5e^{-4}$\\
    \bottomrule
  \end{tabular}
  \label{tab:addlabel}%
\end{table}%

\begin{table}[]
\caption{Task 1 (NIH Chest x-ray diagnosis): Statistics of positive samples among different clients for the experiments that simulate the label set mismatch problem. Each client has 3 locally identified classes and images can be partially labeled.} \label{tab:2}
\centering
\begin{tabular}{@{}c|cccccccc@{}}
\toprule

\multirow{2}{*}{Disease} &\multicolumn{8}{c}{Devices}\\ \cmidrule{2-9}
& A & B & C & D & E & F & G & H\\ \midrule
Consolidation & 309 & 0 & 0 & 0 & 0 & 0 & 0 & 297\\ 
Pneumonia  & 0 & 88 & 0 & 0 & 0 & 0 & 0 & 0 \\
Effusion  & 0 & 0 & 0 & 951 & 0 & 0 & 0 & 0\\
Emphysema & 0 & 0 & 0 & 0 & 161 & 0 & 0 & 0 \\
Edema & 150 & 0 & 0 & 0 & 0 & 0 & 0 & 0  \\
Atelectasis & 0 & 0 & 0 & 918   & 0 & 0 & 0 & 0  \\
Nodule & 0 & 0 & 0 & 0 & 536 & 505 & 0 & 0 \\
Mass & 0 & 0 & 447 & 0 & 0 & 0 & 0 & 0  \\
Thickening  & 0 & 0 & 229 & 0 & 0 & 231 & 0 & 0\\
Cardiomegaly & 0 & 0 & 0 & 182 & 0 & 0 & 174 & 0 \\
Fibrosis & 0 & 141 & 0 & 0 & 0 & 139 & 0 & 0 \\
Hernia  & 0 & 14 & 0 & 0 & 0 & 0 & 20 & 15 \\
Pneumothorax & 271  & 0 & 0  & 0 & 296 & 0 & 0 & 0  \\
Infiltration & 0 & 0 & 1512 & 0 & 0 & 0 & 1469 & 1534\\ \midrule
\bottomrule
\end{tabular}
\end{table}

\begin{table}[t!]
  \centering
\caption{Task 2 (ISIC2018 skin lesion diagnosis): Statistics of positive samples among different clients for the experiments that simulate the label set mismatch problem. Each client has 3 locally identified classes.} \label{tab:2}  \begin{tabular}{@{}c|ccccc@{}}
    \toprule
   \multirow{3}{*}{Disease}  &\multicolumn{5}{c}{Devices}\\ \cmidrule{2-6}
      &A &B &C &D &E  \\ \midrule
    MEL & 0 & 0 & 156 & 0 & 0 \\
    NV  & 919 & 915 & 0 & 0 & 0\\
    BCC & 0 & 70 & 80 & 68 & 76\\
    AKIEC & 0 & 0 & 0 & 41 & 41\\
    BKL & 150 & 0 & 0 & 0 & 138 \\
    DF & 0 & 16 & 16 & 13 & 0\\
    VASC & 25 & 0 & 0 & 0 & 0\\
    \bottomrule
  \end{tabular}
  \label{tab:addlabel}%
\end{table}%
\end{document}